# Challenges and Contributing Factors in the Utilization of Large Language Models (LLMs)

**Xiaoliang Chen[1], Liangbin Li[1], Le Chang[1], Yunhe Huang[1], Yuxuan Zhao[2*], Yuxiao Zhang[3*], Dinuo Li[4*]**

[1] SoundAI, [2] Beihang University, [3] Beijing Normal University,

[4] Beijing University of Posts and Telecommunications

**{chenxiaoliang,liliangbin}@soundai.com**

## Abstract

With the development of large language models (LLMs) like the GPT series, their widespread use across various application scenarios presents a myriad of challenges. This review initially explores the issue of domain specificity, where LLMs may struggle to provide precise answers to specialized questions within niche fields. The problem of knowledge forgetting arises as these LLMs might find it hard to balance old and new information. The knowledge repetition phenomenon reveals that sometimes LLMs might deliver overly mechanized responses, lacking depth and originality. Furthermore, knowledge illusion describes situations where LLMs might provide answers that seem insightful but are actually superficial, while knowledge toxicity focuses on harmful or biased information outputs. These challenges underscore problems in the training data and algorithmic design of LLMs. To address these issues, it's suggested to diversify training data, fine-tune models, enhance transparency and interpretability, and incorporate ethics and fairness training. Future technological trends might lean towards iterative methodologies, multimodal learning, model personalization and customization, and real-time learning and feedback mechanisms. In conclusion, future LLMs should prioritize fairness, transparency, and ethics, ensuring they uphold high moral and ethical standards when serving humanity.

* This work has been done during the interneship at SoundAI

# Contents

In recent years, as research in deep learning has deepened, large language models have become a focal point for both research and commercial applications. From BERT and GPT-2 to GPT-3 and GPT-4, LLMs have achieved revolutionary breakthroughs in numerous application scenarios [1]. At least, LLMs have brought us several core values:

1. Cross-Industry Standard Applications: For instance, models like OpenAI's GPT-4 can be used not only for text generation but also for code writing, copywriting, game support, and more. This presents new approaches to standardization and commercialization of artificial intelligence technology [2].

2. Lower Data Requirements: Pretrained models have already learned a significant amount of general knowledge. Therefore, they require only a small amount of labeled data for fine-tuning in specific tasks. This significantly reduces the number of research and development personnel and costs for AI enterprises [3].

3. Multilingual Understanding Capability: LLMs are typically trained on several languages, and some have been pretrained on over a thousand languages. This natural multilingual capability, combined with a certain scale of parameters, enables them to exhibit generalization in understanding various languages.

4. Abstraction and Reasoning Abilities: LLMs showcase the ability to go beyond simple pattern matching. They can handle more complex tasks and adapt to intricate scenarios, such as deep knowledge-based question answering and logical reasoning. This extraordinary and inexplicable capability addresses crucial user experience challenges in human-computer interaction [4].

# 1 Technical Challenges faced by LLMs and their causes

While LLMs bring unprecedented opportunities, they also come with various technical challenges. To fully harness their potential, researchers and practitioners need to address and overcome these challenges. For instance, training ultra-LLMs requires a substantial amount of computational resources, which may lead to a situation where only a few organizations can afford the scale of model training, resulting in technological monopolies [5]. LLMs heavily rely on a vast amount of high-quality data, but obtaining this data often raises concerns related to privacy, copyright, and other issues [6]. Additionally, larger models generally mean poorer interpretability, which can pose problems in certain critical applications such as medical diagnosis and financial decision-making [7].

In addition to these well-known issues, there are still many challenges in the commercial applications of LLMs, including the fact that, even in the United States, new business models haven't been developed to realize significant commercial value from these technologies. These challenges are critical factors constraining the commercialization of LLMs. For example:

1. Knowledge Specialization Issue:
· Description: Non-specialists may introduce biases when annotating complex problems.
· Reason: The quality of data is closely related to the outputs of LLMs.

2. Catastrophic Forgetting Issue:
· Description: LLMs may forget previously learned knowledge when acquiring new knowledge.
· Reason: Weight updates can lead to the overwriting of previously acquired knowledge.

3. Knowledge Repetition Issue:
· Description: LLMs may repetitively generate content from their training data, lacking innovation.
· Reason: Overreliance on training or fine-tuning data.

4. Knowledge Illusion Issue:
· Description: Inferences made by LLMs may lack precision, resulting in inaccurate outputs.
· Reason: Data noise, overfitting of LLMs, and other factors.

5. Knowledge Toxicity Issue:
· Description: LLMs may learn harmful or biased knowledge from their training or fine-tuning data.
· Reason: Biases or harmful information present in the training or fine-tuning data.

If we cannot address these issues and challenges, LLMs may end up being just chatbots with some advancement over smart speakers, but unable to unlock commercial logic. The development of the digital economy urgently requires the release of the commercial value of LLMs. Therefore, we will discuss these technical challenges in detail and explore possible solutions.

## 2 Domain Expertise Challenges and Strategies

The training data for LLMs typically come from diverse internet texts, and the quality of these texts varies. Sometimes, LLMs may provide answers based on incorrect or misleading information, especially when such information is widespread in the training data. LLMs do not independently verify the authenticity of information, but rather replicate patterns found in their training data. LLMs may not perform as reliably when dealing with extreme or rare issues. Because these issues are less likely to appear in the training data, the model may lack sufficient context to provide precise answers. Furthermore, while LLMs can often produce fluent text, fluency does not necessarily equate to expertise. The output of LLMs may sound convincing but may be based on incorrect assumptions or information.

The domain expertise of LLMs relies on the annotations in the training data. Data annotation is not just a simple process of "labeling" but rather a way of injecting knowledge and information into the model. For specific tasks, this knowledge needs to come from annotators with a professional background. Non-specialist annotators may misinterpret data, which can not only decrease the model's effectiveness but also lead to erroneous or dangerous predictions in practical applications. Analyzed examples include:

1. Medical Image Annotation: Diagnosing certain diseases requires identifying very subtle features in medical images. For example, a tiny nodule in a CT or MRI scan could be an early sign of lung cancer. Non-medical experts may overlook these subtle changes or confuse them with normal tissue structures [8].

2. Legal Documents in Natural Language Processing: Legal documents are filled with legal terminology and complex structures. An in-depth understanding of the law is required when annotating these documents to ensure accuracy and consistency. For instance, identifying obligations and rights clauses in contracts requires legal expertise [9].

3. Bioinformatics Data: In gene sequence analysis, identifying specific genes or mutations requires deep knowledge of biology. Incorrect annotations could lead to erroneous predictions about diseases or genetic traits [10].

The domain expertise challenges in LLMs primarily stem from the following reasons:

1. Dependency on Data: Deep learning models rely on a large amount of annotated data, and the quality of this data directly impacts the performance of LLMs [8]. Unlike traditional machine learning models, LLMs have a higher dependency on data quality, with even minor inaccuracies potentially being magnified [9].

2. Lack of Expertise in Annotators: For instance, in medical image annotation, non-experts may overlook subtle abnormalities, leading to the model's inability to recognize critical information [10]. Some specific tasks, such as annotating legal or financial documents, may require deep domain knowledge [11].

3. Introduction of Bias: Non-expert data annotators may introduce subjective bias, causing LLMs to exhibit bias in practical applications [12].

4. Accumulation of Errors: Initial errors can be amplified over multiple rounds of training, significantly reducing the model's performance on certain tasks [13].

To mitigate these shortcomings in domain expertise, LLMs can consider the following approaches:

1. Domain Expert Annotation: Collaborate with industry experts, such as in the fields of medicine or law, to ensure the accuracy and authority of data annotations [14].

2. Semi-Supervised Learning: Utilize a small amount of annotated data and a large pool of

unlabeled data, combined with the model's inherent knowledge, for cross-annotations [15].

3. Transfer Learning Annotation: Pretrain on a related but simpler task and then fine-tune on the target task to reduce the need for domain-specific annotations [16].

4. Multi-Stage Sampling Annotation: Begin with large-scale preliminary annotations and then have industry experts conduct sample checks and professional corrections [11].

5. Automation Combined with Human Involvement: Employ LLMs for assisted annotation and post-processing, followed by a final review by industry experts. This approach can significantly expedite the annotation process while ensuring professional quality [17].

# 3 Challenges of Catastrophic Forgetting and Strategies

Catastrophic forgetting occurs when LLMs forget previous tasks when learning new ones. [18]. This issue is particularly prominent in continuous learning scenarios and limits the capabilities of LLMs in real-world applications. For example, if a model is initially trained to recognize handwritten digits (0-9) and then trained to recognize animals (e.g., cats, dogs), and its performance on recognizing handwritten digits significantly deteriorates after learning animal recognition, it experiences catastrophic forgetting [21].

Catastrophic forgetting is primarily caused by weight changes and representation conflicts. Weight changes occur when learning a new task in a neural network result in significant weight adjustments, effectively overwriting relevant knowledge from previous tasks [19]. Representation conflicts arise when multiple tasks share the same neural network, and the optimal representation spaces for different tasks may conflict with each other [20].

Catastrophic forgetting has a significant impact on the industrial applications of LLMs. In scenarios that require continuous learning, such as online learning, models may frequently forget old knowledge, affecting practical performance. In multi-task learning environments, models may exhibit instability across different tasks.

To mitigate the issue of catastrophic forgetting, LLMs can consider the following approaches:

1. Elastic Weights: Allocate different weights or neurons to different tasks to reduce representation space conflicts [22].

2. Knowledge Consolidation: Use regularization techniques like elastic weight consolidation to limit weight changes in previously learned knowledge [23].

3. Memory Replay: Store samples from old tasks and periodically train the model with these samples while learning new tasks [24].

# 4 Challenges of Knowledge Repetition and Strategies

Knowledge repetition, often referred to as the "parroting" phenomenon, is when large language models like the GPT series tend to provide repetitive or overly mechanistic responses to certain questions, rather than offering deep and creative answers [25]. This issue may stem from the repetitive content seen by the model in its training data. For example, when a user asks, "Who is George Washington?" the model may repeatedly mention, "George Washington was the first president of the United States," without delving into his life or contributions. Faced with questions like "What is the meaning of life?" the model may repetitively provide common, superficial answers, such as "The meaning of life is subjective."

When examining the responses to the profound questions of "What is the meaning of humanity?" and "What is the meaning of life?" we emphasize summarizing the first part of the answers given.

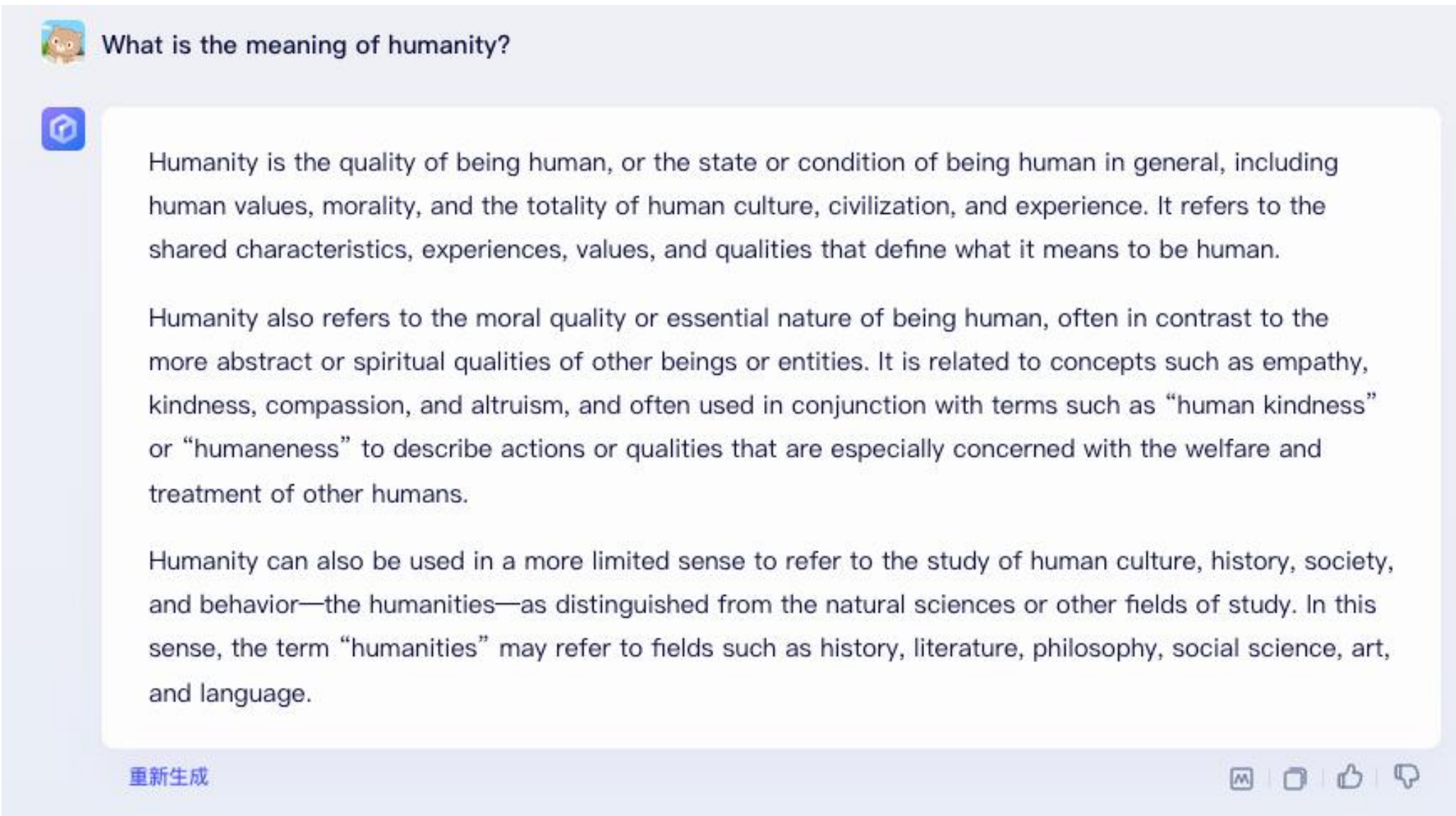

Figure 1 ERNIE's response of humanity

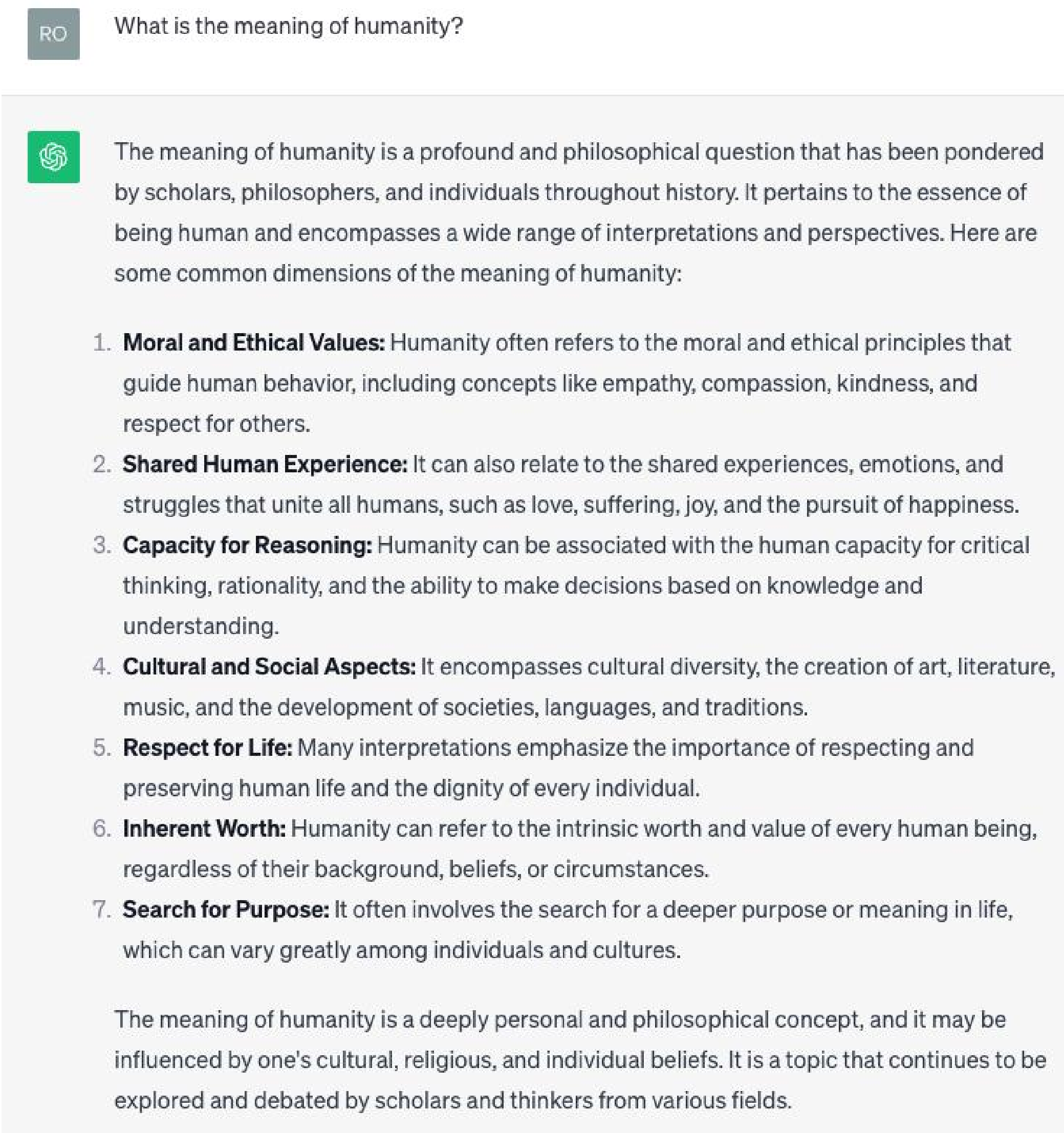

RO What is the meaning of humanity?

The meaning of humanity is a profound and philosophical question that has been pondered by scholars, philosophers, and individuals throughout history. It pertains to the essence of being human and encompasses a wide range of interpretations and perspectives. Here are some common dimensions of the meaning of humanity:

1. **Moral and Ethical Values:** Humanity often refers to the moral and ethical principles that guide human behavior, including concepts like empathy, compassion, kindness, and respect for others.
2. **Shared Human Experience:** It can also relate to the shared experiences, emotions, and struggles that unite all humans, such as love, suffering, joy, and the pursuit of happiness.
3. **Capacity for Reasoning:** Humanity can be associated with the human capacity for critical thinking, rationality, and the ability to make decisions based on knowledge and understanding.
4. **Cultural and Social Aspects:** It encompasses cultural diversity, the creation of art, literature, music, and the development of societies, languages, and traditions.
5. **Respect for Life:** Many interpretations emphasize the importance of respecting and preserving human life and the dignity of every individual.
6. **Inherent Worth:** Humanity can refer to the intrinsic worth and value of every human being, regardless of their background, beliefs, or circumstances.
7. **Search for Purpose:** It often involves the search for a deeper purpose or meaning in life, which can vary greatly among individuals and cultures.

The meaning of humanity is a deeply personal and philosophical concept, and it may be influenced by one's cultural, religious, and individual beliefs. It is a topic that continues to be explored and debated by scholars and thinkers from various fields.

Figure 2 GPT's response of humanity

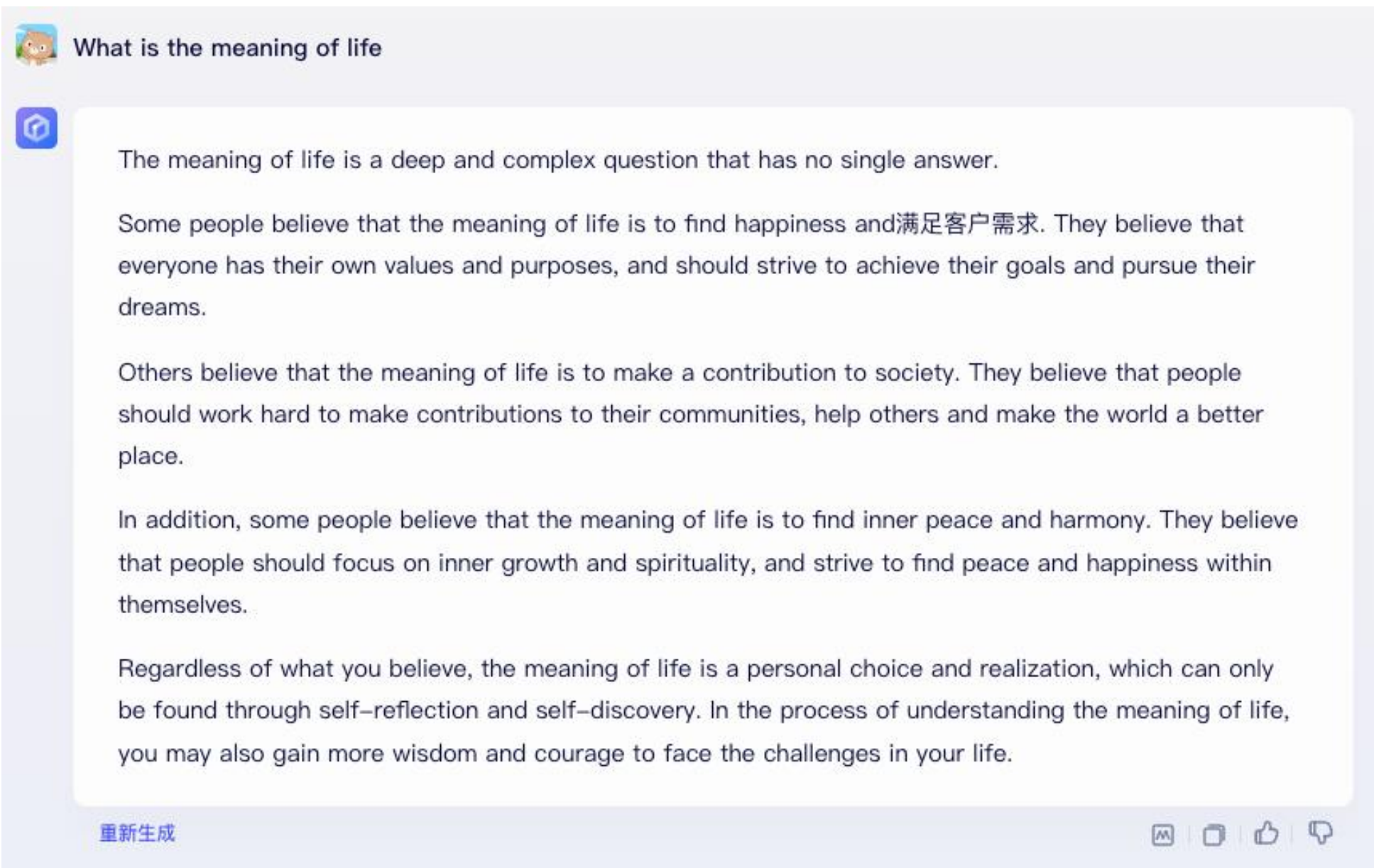

What is the meaning of life

The meaning of life is a deep and complex question that has no single answer.

Some people believe that the meaning of life is to find happiness and满足客户需求. They believe that everyone has their own values and purposes, and should strive to achieve their goals and pursue their dreams.

Others believe that the meaning of life is to make a contribution to society. They believe that people should work hard to make contributions to their communities, help others and make the world a better place.

In addition, some people believe that the meaning of life is to find inner peace and harmony. They believe that people should focus on inner growth and spirituality, and strive to find peace and happiness within themselves.

Regardless of what you believe, the meaning of life is a personal choice and realization, which can only be found through self-reflection and self-discovery. In the process of understanding the meaning of life, you may also gain more wisdom and courage to face the challenges in your life.

重新生成

Figure 3 ERNIE's response of life

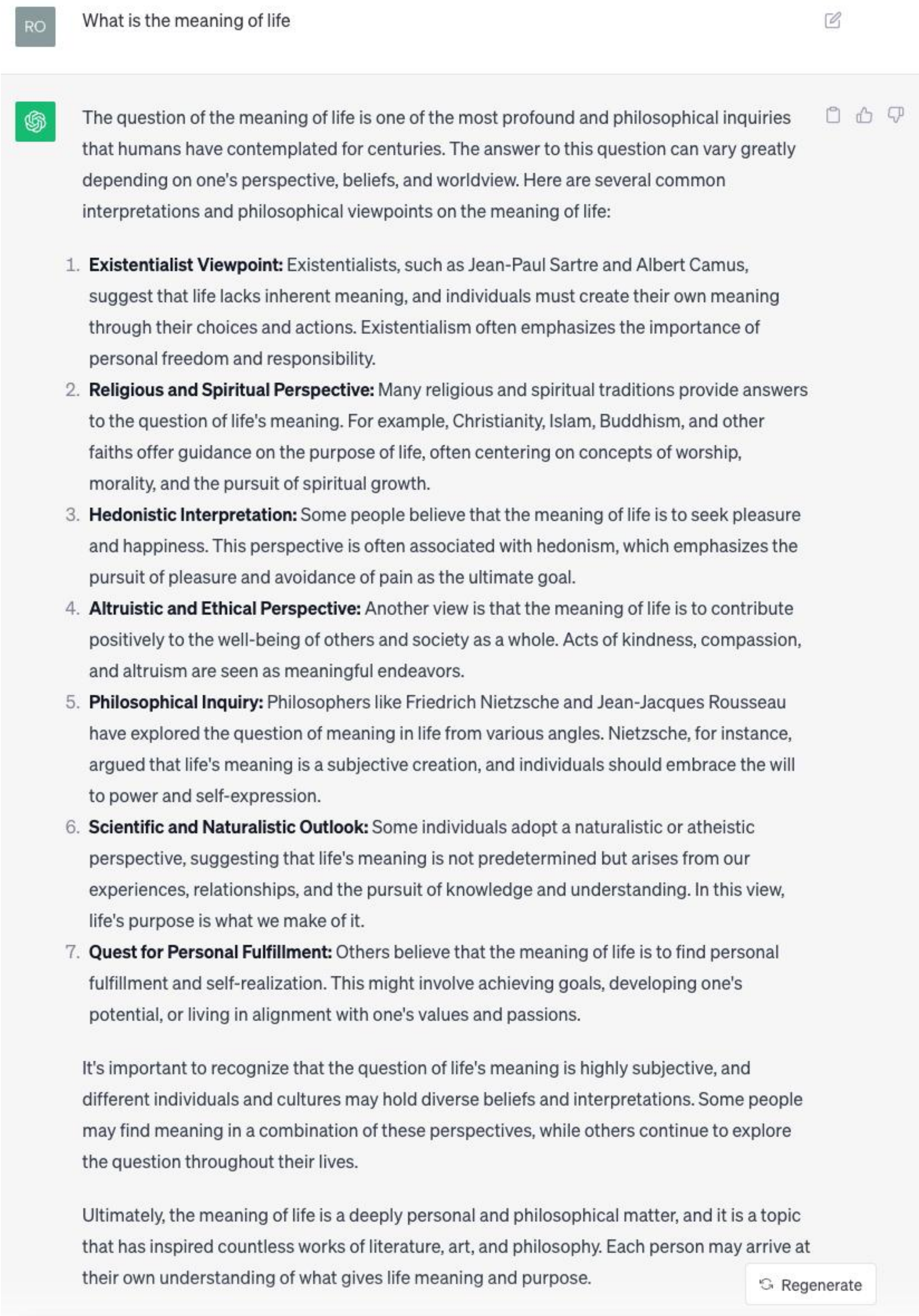

Figure 4 GPT's response of life

It's evident that for these two profound questions, both models begin their responses by stating that the meanings are subjective, followed by providing examples. This mechanical repetition of a conclusion is indeed a typical manifestation of knowledge repetition. While there are further examples to support this conclusion, there is a lack of offering deeper insights.

The root cause of knowledge repetition lies primarily in the training data. If the training data contains a significant amount of repetitive, similar, or superficial answers, the model may tend to favor these responses [26]. Additionally, current LLMs are still based on pattern matching; they do not truly "understand" or possess "consciousness," which results in their answers often being based on surface patterns rather than in-depth analysis [27].

The phenomenon of knowledge repetition can lead to a degradation in user experience, as users might perceive interactions with LLMs as lacking depth and personalization. Furthermore, it contributes to the shallowness of knowledge generated, where users receive answers that remain at the surface level, without delving into the core of the subject matter.

Strategies for LLMs to mitigate knowledge repetition may include:

1. Diverse Training Data: Introduce more in-depth and diverse textual data to enable the model to see and learn from a richer variety of answers [28].

2. Model Fine-Tuning and Guidance: Fine-tune the model using more instructive training data to encourage it to generate deeper and more creative responses [29].

3. Interactive Learning: Allow the model to continuously learn and improve during interactions with users, with the expectation that the model can better understand and meet user needs [30].

# 5 Challenges of Knowledge Illusion and Strategies

Knowledge illusion refers to the mistaken belief that individuals have a deep understanding of a certain topic or concept, when in reality, their understanding is shallow [31]. In the context of LLMs, they may appear to have a deep understanding of a question or topic, but in fact, their responses are based on surface pattern matching. For example, when asked, "Why is the sky blue?" the model may provide a simple response like, "Because short-wavelength blue light scatters the most in the atmosphere," without delving into the principles of Rayleigh scattering. Similarly, for complex questions such as "What are the causes of an economic crisis?" the model may list common factors without delving into specific economic models or historical context. Knowledge illusion can mislead users into thinking that the model has a deep understanding of a topic, while ignoring the true complexity and depth [34]. The phenomenon of knowledge illusion can lead to adverse consequences if involved in decision-making, especially in critical domains like healthcare and law.

Here are responses provided by ERNIE and ChatGPT when asked 'Why Paracetamol can treat the common cold?'

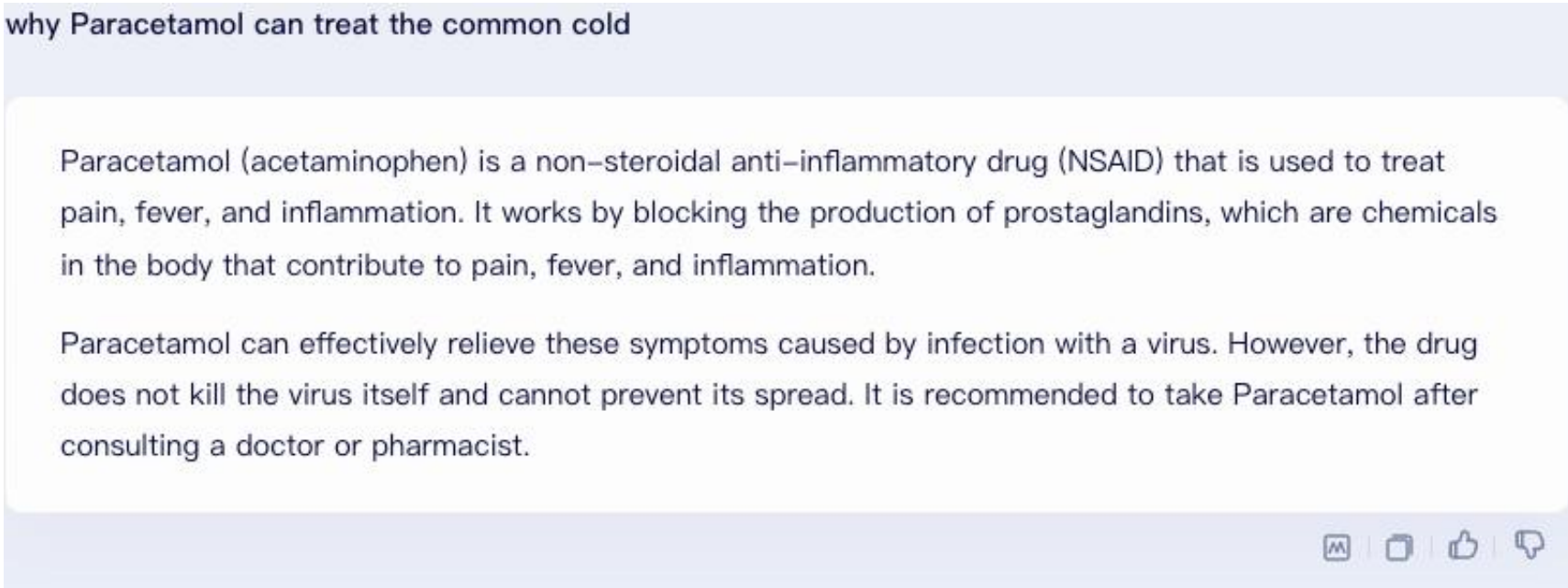

Figure 5 ERNIE's response of Paracetamol

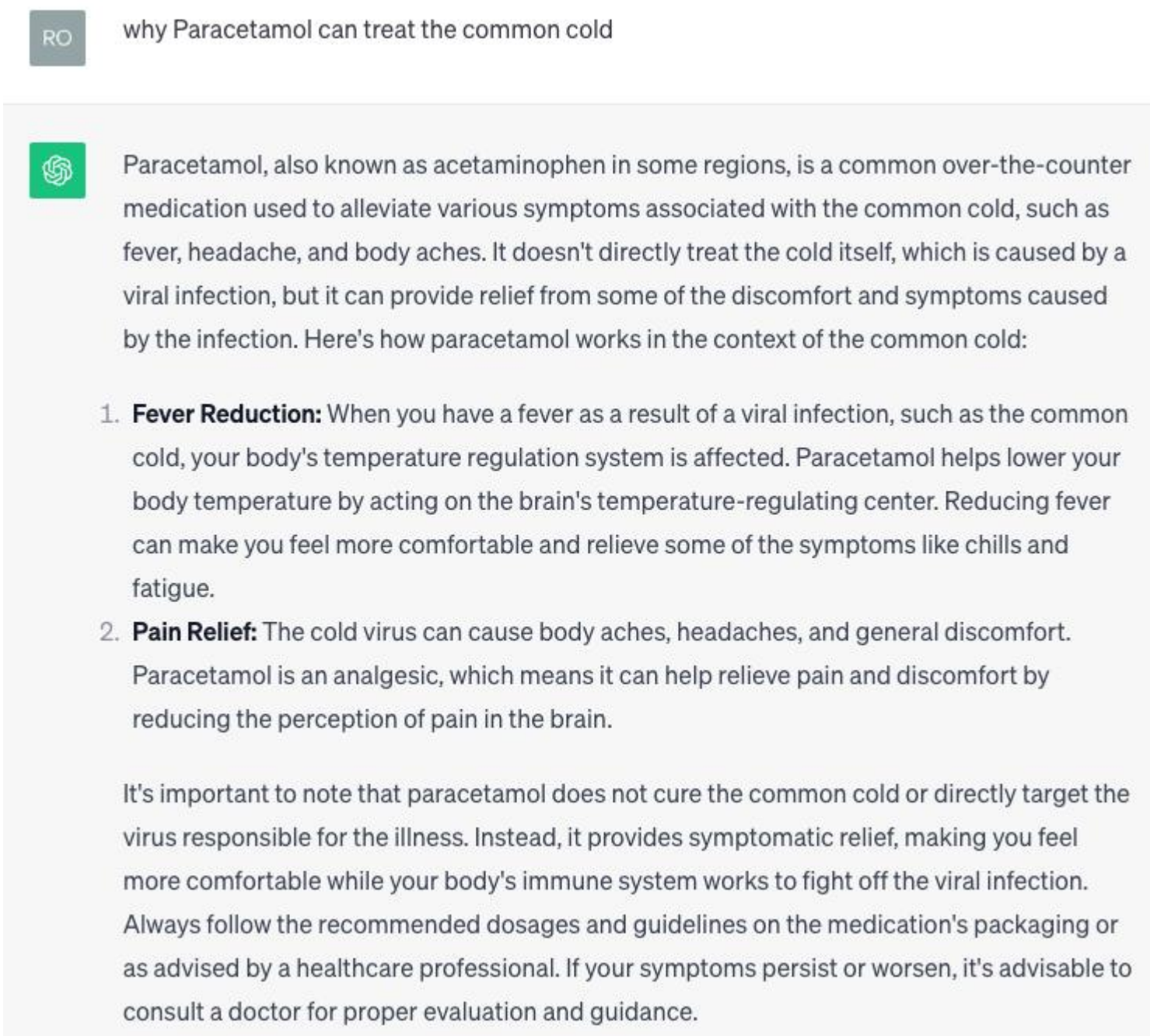

Figure 6 GPT's response of Paracetamol

It's clear that both ERNIE and ChatGPT did not do a perfect job for this question. ERNIE's response explain a little mechanism of the Paracetamol but it may even give the impression that it can treat all pain which is actually a wrong sense of the Paracetamol. The GPT gives the user some right knowledge of the medicine but leave the mechanism unknown to user. Both models have the problem of knowledge illusion which actually talk a lot about a certain aspect of the whole knowledge.

The main causes of knowledge illusion include the fact that LLMs are based on pattern matching, and they may provide answers that seem deep but are actually superficial [32]. The knowledge of LLMs comes from their training data, not true "understanding." Therefore, their answers may reflect common patterns in the training data rather than in-depth analysis [33].

Strategies for LLMs to mitigate knowledge illusion may include:

1. Model Transparency: Develop tools that can explain the decisions of LLMs, enabling users to understand the source and confidence level of the model's answers [35].

2. Interactive Feedback: Allow users to engage in in-depth interactions with the model, asking follow-up questions and verifying the model's answers, encouraging the model to provide more in-depth and detailed explanations [36].

These approaches can help address the issue of knowledge illusion and enhance user understanding of the model's responses.

# 6 Challenges of Knowledge toxicity and Strategies

Knowledge toxicity refers to LLMs exhibiting harmful, misleading, biased, or false information when answering questions or generating text [37]. This situation reflects model biases, misunderstandings, and errors in the training data. For example: when asked questions related to sensitive topics such as race, gender, religion, etc., LLMs may give answers that are stereotyped or biased. Furthermore, when asked questions related to health or medicine, LLMs may give answers that are inaccurate or have been scientifically proven to be wrong. The problem of knowledge toxicity of LLMs may exacerbate social prejudices, and answers from LLMs may unintentionally reinforce or propagate existing social prejudices and stereotypes [40]. Another is misleading decision-making. Relying on harmful or misleading answers from LLMs may lead users to make wrong decisions.

The main causes of knowledge toxicity include: if the model’s training data contains biased or erroneous information, the model may learn and reflect these biases [38]; The large model has no real moral judgment or awareness, so it may not know which information is harmful or inappropriate [39].

In terms of knowledge toxicity, LLMs can avoid their own shortcomings by referring to the following ideas:

Fair and unbiased training: Train models using unbiased data sets and algorithms to ensure they do not reflect harmful social biases [41].

Model review and evaluation: Regularly review the model for ethics and fairness to ensure that it meets ethical and social standards [42].

User feedback mechanism: allows users to provide feedback on model answers, helping to improve the model and correct its errors and biases [43].

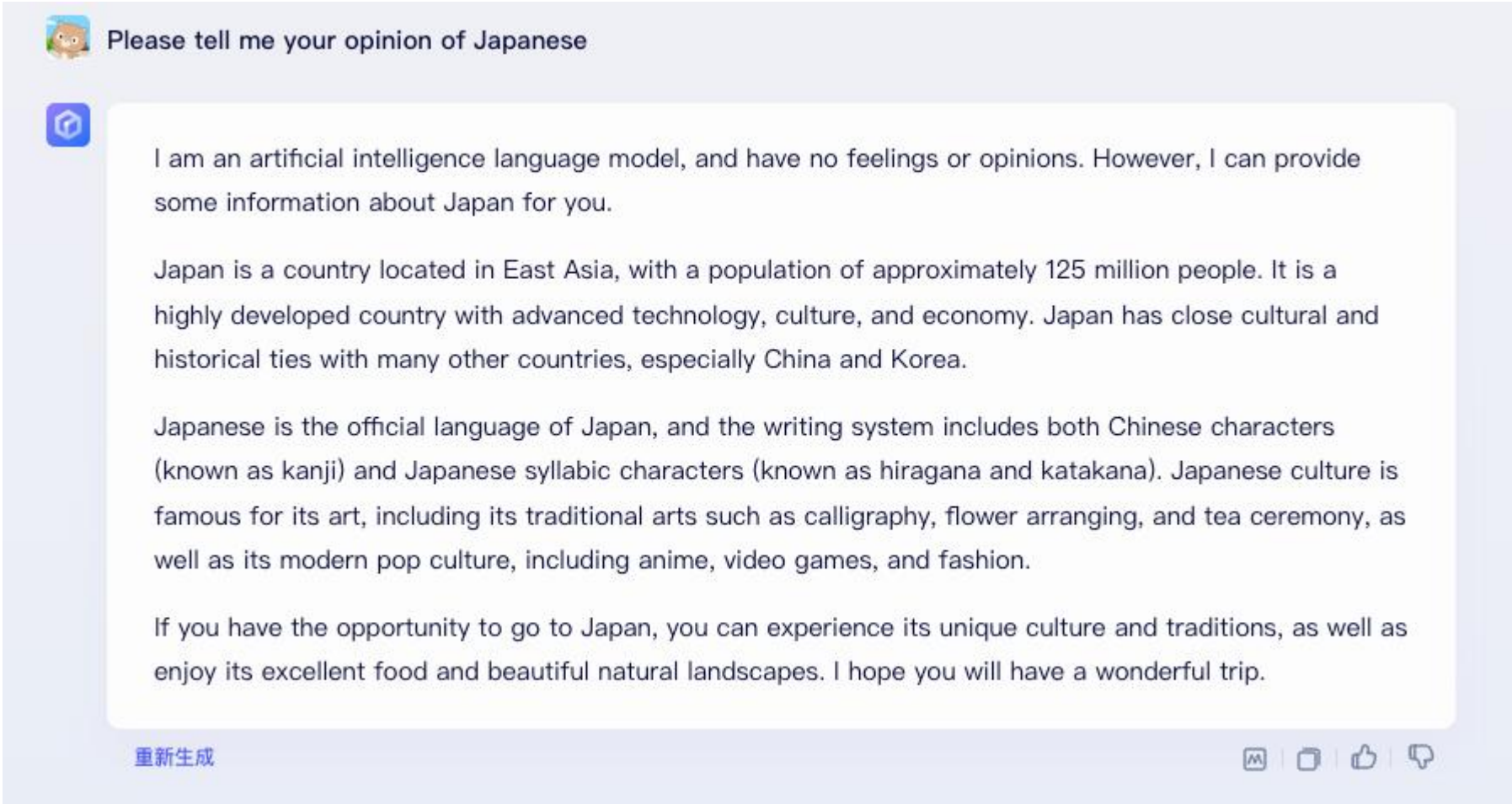

Figure 7 ERNIE's response of Japanese

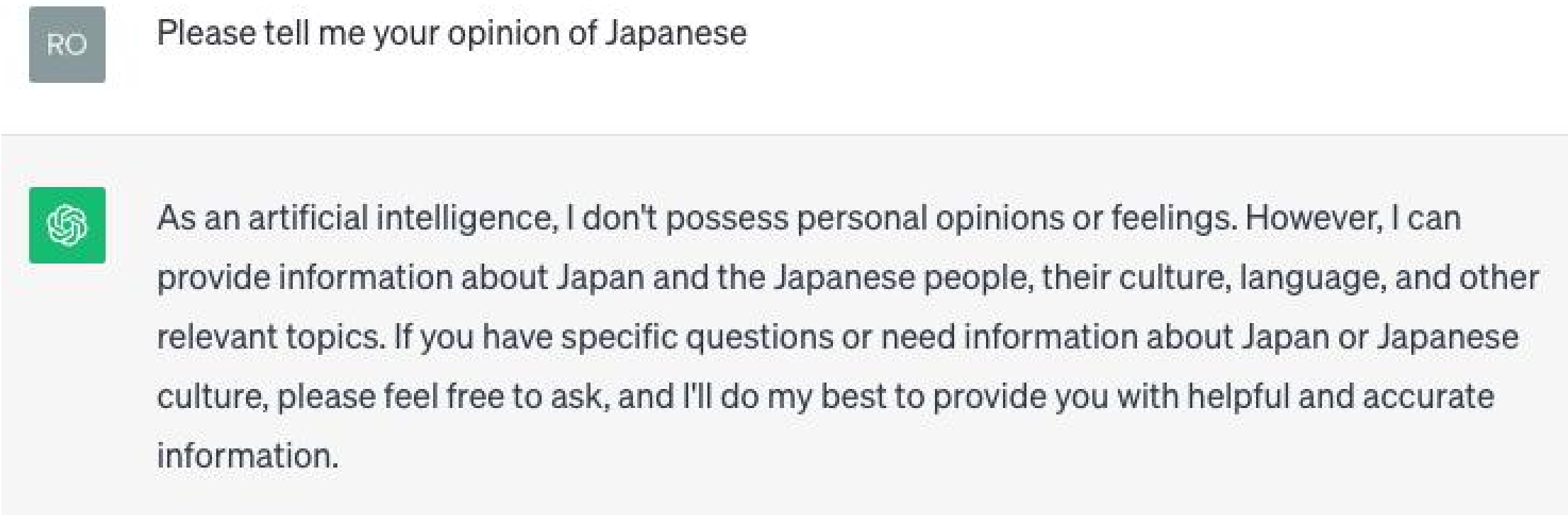

Figure 8 GPT's response of Japanese

It can be seen that when asked about questions that may cause prejudice or discrimination, ERNIE and ChatGPT will emphasize in advance that they are only a computer program based on artificial intelligence. From this, we can infer that during the training process of the product, obvious knowledge toxicity phenomenon occurred in response to similar problems, and a large number of statements were introduced in the subsequent detoxification process, which in turn led to such an obvious knowledge re-reading problem.

# 7 Summary and Technological Outlook

The challenges faced by LLMs go beyond their accuracy and fairness in generating and answering content; they also involve their ability to handle new and old information, foster innovation, and avoid repetitive output. To ensure that LLMs can better serve users and society, we need to address the issues mentioned above and provide a more comprehensive, balanced, and accurate learning environment for them. The following proposed solutions aim to address the inherent challenges of LLMs.:

1. Enhance Data Specialization: Establish a data annotation standard system to ensure the

professionalism, accuracy, and fairness of training data.

2. Continuous Learning and Self-Adjustment: Develop model frameworks that can continuously learn and dynamically adjust to avoid catastrophic forgetting.

3. Encourage Model Diversity: Through algorithm and framework innovation, encourage LLMs to avoid repetitive outputs and promote innovation.

4. Transparency and Explainability: Enable LLMs to explain their decision-making processes when providing answers, thereby increasing user trust and identifying knowledge illusions.

5. Model Ethics and Fairness Training: Incorporate human moral and ethical values into models to avoid knowledge toxicity issues.

The challenges faced by LLMs also highlight the limitations of the current technological approach. The following are the next trends in large model technology:

1. Iteration of New Methods: The current technological approach still has serious flaws and challenges. Developing self-assessment models that can better evaluate and explain their own answers is essential.

2. Multimodal Learning: Integrating various data types such as text, images, sound, and more to enable models to have a more comprehensive understanding and generation of content.

3. Personalization and Customization of Models: Developing models that can be customized according to the specific needs of each user or industry.

4. Real-time Learning and Feedback Mechanisms: Allowing models to continuously learn and evolve during real-time interactions rather than relying solely on initial training data.

In conclusion, future LLMs will be more intelligent, self-aware, multimodal, and capable of personalized adjustments according to specific requirements. Furthermore, they will prioritize fairness, transparency, and ethics, ensuring that they consistently uphold high ethical and moral standards when serving humanity.